\documentclass[10pt,twocolumn,letterpaper,conference]{IEEEtran}
\IEEEoverridecommandlockouts

\usepackage{cite}
\usepackage{amsmath,amssymb,amsfonts}
\usepackage{algorithmic}
\usepackage{graphicx}
\usepackage{textcomp}
\usepackage{xcolor}
\usepackage{lettrine}
\def\BibTeX{{\rm B\kern-.05em{\sc i\kern-.025em b}\kern-.08em
    T\kern-.1667em\lower.7ex\hbox{E}\kern-.125emX}}
    
\usepackage[ruled]{algorithm2e}
\renewcommand{\algorithmcfname}{ALGORITHM}
\SetAlFnt{\small}
\SetAlCapFnt{\small}
\SetAlCapNameFnt{\small}
\SetAlCapHSkip{0pt}
\IncMargin{-\parindent}

\usepackage{moreverb,algorithmic,subfigure}
\usepackage{amsmath}
\usepackage{amssymb,lscape}
\usepackage{tabularx}
\usepackage{balance}
\usepackage{times}
\usepackage{float}

\newcommand{\be}{\begin{equation}}
\newcommand{\ee}{\end{equation}}

\usepackage{epsfig}
\usepackage{graphicx}
\usepackage[figuresright]{rotating}
\usepackage{subfigure}
\usepackage{amssymb}
\usepackage{amsmath}
\usepackage{setspace}
\usepackage{rotating}
\usepackage{multirow}
\usepackage{wrapfig}
\usepackage{booktabs}
\usepackage{array}
\usepackage{comment}
\usepackage{subfigure}
\usepackage{epstopdf}
\usepackage{epsfig}
\usepackage{amsmath}
\usepackage{amssymb}
\usepackage{times}

\usepackage{textcomp}

\usepackage[ruled]{algorithm2e}
\renewcommand{\algorithmcfname}{ALGORITHM}
\SetAlFnt{\small}
\SetAlCapFnt{\small}
\SetAlCapNameFnt{\small}
\SetAlCapHSkip{0pt}
\IncMargin{-\parindent}

\usepackage{varwidth,xcolor}
\usepackage{graphicx}

\usepackage{moreverb,algorithmic,subfigure}
\usepackage{amsmath}
\usepackage{amssymb,lscape}
\usepackage{tabularx}
\usepackage{textcomp}
\usepackage[normalem]{ulem}
\useunder{\uline}{\ul}{}

\usepackage{colortbl}

\begin{document}


\title{\huge{Source Feature Compression for Object Classification in Vision-Based Underwater Robotics}}


\author{\IEEEauthorblockN{Xueyuan Zhao,~\IEEEmembership{Member,~IEEE}, Mehdi Rahmati,~\IEEEmembership{Senior Member,~IEEE}, \\Dario Pompili,~\IEEEmembership{Fellow,~IEEE}}
\thanks{\IEEEcompsocthanksitem
X.~Zhao and D.~Pompili are with the Dept. of Electrical and Computer Engineering~(ECE), Rutgers University, NJ. M.~Rahmati is with the Dept. of Electrical Engineering and Computer Science, Cleveland State University, OH; he worked on this project as a graduate student while at Rutgers. \protect 
\\E-mails: \{xueyuan.zhao, mehdi.rahmati, pompili\}@rutgers.edu. \protect
\IEEEcompsocthanksitem {This work was supported by the NSF NeTS Award No.~CNS-1319945.}}
}
\markboth{Submitted to IEEE Transactions on Pattern Analysis and Machine Intelligence~(TPAMI), October 2021}
{Shell \MakeLowercase{\textit{et al.}}: Bare Demo of IEEEtran.cls for Journals}

\clearpage
\maketitle

\thispagestyle{empty}
\pagenumbering{arabic}

\begin{abstract}
New efficient source feature compression solutions are proposed based on a two-stage Walsh-Hadamard Transform~(WHT) for Convolutional Neural Network~(CNN)-based object classification in underwater robotics. The object images are firstly transformed by WHT following a two-stage process. The transform-domain tensors have large values concentrated in the upper left corner of the matrices in the RGB channels. By observing this property, the transform-domain matrix is partitioned into \textit{inner} and \textit{outer} regions. Consequently, two novel partitioning methods are proposed in this work: $(i)$~fixing the size of inner and outer regions; and $(ii)$~adjusting the size of inner and outer regions adaptively per image. The proposals are evaluated with an underwater object dataset captured from the Raritan River in New Jersey, USA. It is demonstrated and verified that the proposals reduce the training time effectively for learning-based underwater object classification task and increase the accuracy compared with the competing methods. The object classification is an essential part of a vision-based underwater robot that can sense the environment and navigate autonomously. Therefore, the proposed method is well-suited for efficient computer vision-based tasks in underwater robotics applications. 
\end{abstract}

\section{Introduction}\label{sec:intro}

\lettrine{D}{eployment} of computer vision and machine learning algorithms in Autonomous Underwater Vehicles~(AUVs) is a challenging task due to the many existing constraints in the underwater environment, as underwater robots become pervasive in applications such as oceanographic data collection, pollution and environmental monitoring, tsunami detection/disaster prevention, and assisted navigation. Underwater robots are normally battery-powered with limited on-board computing, memory, and storage capabilities. Therefore, to deploy Convolutional Neural Network~(CNN)---which is widely adopted in computer vision for its performance---the technical challenges of performing efficient CNN training in underwater robotics must be addressed. The image of objects obtained from the underwater environment are different from the ones in existing commonly accepted image datasets for CNN training (e.g., ImageNet, CIFAR-10 datasets). These underwater objects include rocks and wood branches that naturally exist at the floor of rivers, lakes or oceans. The captured images of these objects are distorted due to the light scattering and absorption, which leads to the lack of contrast, blurry appearance, and distorted color~\cite{chen2019real}. For these reasons, pre-trained CNNs based on commonly-accepted image databases are not applicable to the underwater vision-based tasks; therefore, training of the images needs to be done efficiently with the collected underwater images. 

The training of CNN architecture is known to be demanding in terms of computation, power, memory, and storage. In practice, underwater robots is sent to nagivate in the underwater environment to perform underwater video and image data collection task in the first stage. The collected video and image datasets in the current river/ocean environments are applied to perform the training of the underwater robots based on the computer vision CNN structure. Due to differences in the underwater environments, the object shape and image color are different in a particular environment. To adapt the CNN architecture to the data, the training with the collected datasets in a particular environment is very necessary. The trained CNN is then applied to real-time computer-vision inference in subsequent underwater navigation tasks. Thus, it is expected that the training time is relatively short so that the robot can proceed with the autonomous navigation shortly after the video/image collection task. Moreover, the online object classification inference in underwater robots should be performed with high power efficiency as well as with low latency, while saving memory and storage utilization.

There is WHT-based CNN structure~\cite{Deveci18} research work for image classification with binary weightings in the CNN structure. The DCT is studied on the CNN structure for efficient computer-vision tasks~\cite{Verma18}. In these existing works, i.e., \cite{Deveci18,Verma18}, the operations are all linear transformations. For example, in the WHT-based CNN structure~\cite{Deveci18}, the transformation of WHT is a pure linear operation without any non-linear operation. Compared with the existing WHT-based method, in our proposal the non-linear operations are designed to partition the post-WHT datasets into \emph{inner} and \emph{outer regions}, and the non-linear max-pooling techniques are applied to these partitioned regions. This non-linear operation has several benefits in reducing the training speed while improving the classification accuracy. The proposal is designed to process the underwater image data and is evaluated and compared against the WHT-based and the DCT-based approaches.

\textbf{Our Methods:}
In this work, new source feature compression methods are proposed to address the above challenges in deploying computer vision methods in autonomous underwater vehicles. A two-stage Walsh-Hadamard Transform~(WHT) is initially performed by a column-vector WHT, then followed by a row-vector WHT, both in RGB channels. It is observed that large values are concentrated in the upper left corner of two-stage WHT matrices. Based on this observation, new source compression methods are proposed. In the first proposed method, the WHT matrices are partitioned into the \emph{inner} and \emph{outer regions} with fixed sizes. In the inner region of the upper left corner of the matrix, the max-pooling is done only on three RGB channels; conversely, in the outer region, the max-pooling is performed on matrices or vectors on three RGB channels. The resulting tensors are padded with zero in the null areas. The proposal has the advantage of reducing the size of outer regions while keeping the relevant information in the two-stage-WHT tensor. The reason to choose WHT is that the transform-domain vector has the information of the vector extracted in the low-dimension which is very helpful in designing the image source compression scheme. In the second proposed method, the sizes of the inner and outer regions are adjusted adaptively in each image. The size adjustments are determined via comparison of an image-specific metric with a predefined threshold. By testing with the underwater datasets, the computation time is significantly reduced; in the meanwhile, the proposal has a considerable accuracy with respect to the similar methods such as WHT-only method and Discrete Cosine Transform~(DCT)-based methods. These benefits are indicated by the experiments that test the performance of the proposal and the competing methods on the collected underwater object image dataset. 

{
Our collected underwater object image dataset has differences from the widely accepted dataset of ImageNet. The ImageNet is a general dataset that has a wide range of objects found in the daily life, while our dataset is a specialized dataset that only contains the objects found in underwater videos recorded by underwater robots. In our datasets, we only have 6 classes of objects, including large and small stones, large and small woods. While in ImageNet, there are a large number of classes of objects: more than 21 thousand classes of manually labeled images are collected in the ImageNet database. A subset of ImageNet named Tiny ImageNet has 200 classes of images. In addition, the color diversity in our underwater image is also much less than the color diversity in the ImageNet dataset. Our dataset is a specialized dataset that has fewer classes of objects and less color diversity compared with the ImageNet dataset. The proposed method in the work is particularly designed for the underwater image datasets.}

Our work is the first proposal of non-linear regional max-pooling operation for source compression in the transform-domain tensors. \textit{The major novelty of our work is to introduce this non-linear operation into the source feature compression for CNN-architecture-based image classification.} This regional max-pooling operation has the technical merit of reducing the training complexity and training time and, at the same time, it removes the noise from source images resulting in machine learning classification accuracy improvement.

\textbf{Our Contributions:}
New source feature compression methods are proposed in this work to reduce training time and computational complexity compared to competing methods such as WHT-only method and DCT method. Since the transform-domain tensors have information concentrated in low dimension, the max-pooling operation in outer regions can possess the effect of removing noise in the original images. The proposed solution is well-suited for the applications in robotics where there are constraints on power consumption, memory, storage, and computing capabilities. To verify the effectiveness of our proposal, the image datasets are captured from the bottom of the Raritan River in New Jersey, USA. The CNN training of the proposal is tested with the robotic measured data and is shown to significantly reduce training time in classifying underwater objects including rocks and wood branches. Meanwhile, the accuracy is improved w.r.t. the methods under comparison including WHT- and DCT-based methods. The adaptive region proposal further enhances the convergence performance with the optimized CNN architecture. Our contributions can be summarized as follows:
\begin{itemize}
     \item New source compression methods are proposed for CNN-based underwater object classification. The proposals are based on the partitioning of the tensor after a two-stage Walsh-Hadamard Transformation~(WHT) and a max-pooling compression on the partitioned data matrices. The first proposal works by fixing the inner and outer region on the transform-domain matrices after two-stage WHT.
     
     \item An adaptive region method is further proposed based on the two-stage WHT transformation with adaptive inner and outer region partitioning. The sizes of the inner and outer regions are adjusted adaptively per image to improve the performance of the object classifier compared with the first fixed-region proposal.
    
     \item The implementations of the proposed solutions are verified with underwater image datasets collected in underwater experiments including rocks and wood branches to perform underwater object classification. The training time of the proposal can be reduced by several folds while improved classification accuracy is achieved compared to the existing methods.

\end{itemize}

In this work, experiments are performed on the underwater scene images captured in the river environment. The dual advantages of reducing the computational time while improving the accuracy are verified via experiments. A basic convolutional neural network architecture is adopted in the evaluation and its architecture parameters are optimized on both the fixed-region and the adaptive-region proposals. The results include the classification accuracy convergence curve with varied parameter settings on both proposals as well as two competing existing methods, which are outperformed by our new methods.

\textbf{Article Outline:}
The rest of paper is organized as follows. In Sect.~\ref{proposal}, the descriptions of our proposed solutions including the fixed-region proposal and adaptive region proposal are provided. In Sect.~\ref{performance}, the proposals are evaluated with the underwater datasets and the results are compared with the existing approaches. The conclusions are drawn in Sect.~\ref{conclusions}.

\section{Related Work}\label{related_work}
We present related work in the recent literature on underwater object classification and CNN in vision applications.

\textbf{Underwater Object Classification:}
Object classification is one of the most challenging tasks in the underwater environment. Two types of common sensors, which are widely used in underwater vehicles, are sonar and vision cameras~\cite{Rahman18}. The images taken by the sonar cameras depict the floor are useful for bathymetry estimation, even in murky water with low visibility. However, the sonar camera usually cannot record and extract the details and the features of the images of the underwater objects. On the other hand, visual cameras of underwater objects, providing more features, are the imaging choice for object detection and classification applications such as in the Simultaneous Localization and Mapping~(SLAM) technique~\cite{rahmatislam2018}. Many factors affect the quality of underwater images including absorption, scattering, and color distortion. Therefore, imaging models are required to compensate for these factors such as the one proposed in~\cite{lu2015contrast} to combat attenuation discrepancy along the propagation path. Although the visual cameras mounted on the AUVs can capture large amounts of video data, there are bottlenecks in transmitting the video data in real-time if the AUV is connected to the control center via a wireless link. The acoustic signal can propagate for kilometers of distance, but since the bandwidth is narrow and is mostly allocated for controlling and commanding needs, the object classification tasks are expected to be performed by on-board computing resources on the robot.

\textbf{CNN in Vision Applications:}
The typical CNN architectures~\cite{Alom18} adopted in computer vision include AlexNet, VGG, ResNet, Inception, Xception, MobileNetV1 and MobileNetV2. The AlexNet architecture~\cite{Alex12} is a well-accepted basic CNN architecture composed of five convolutional layers and three FC layers. In the VGG architecture~\cite{Karen14}, there are five blocks, and each block is composed of 2-3 convolutional layers and one max-pooling layer, with two FC-layer at the end of the network. The ResNet architecture~\cite{He16} has a much deeper structure of fifty convolutional layers in the original proposal. There are Inception architecture~\cite{Szegedy15} and Xception architecture~\cite{Chollet16} proposals and the Xception architecture outperforms the competing VGG, ResNet on the ImageNet datasets. The MobileNetV1~\cite{Howard17} and MobileNetV2~\cite{Sandler18} are the recent proposals on CNN for mobile computing devices with reduced complexity compared to other architectures. In this work, we are adopting a basic CNN architecture to evaluate the proposal. The reason to adopt the basic architecture instead of advanced existing architecture is that, we are comparing our proposal with the competing proposal under the same condition. The above architectures are mostly optimized in general public datasets that are composed of objects in daily life including ImageNet datasets. 

\section{Proposed Solutions}\label{proposal}

 \begin{figure}
 \begin{center}
 \scalebox{0.12}
 {\includegraphics{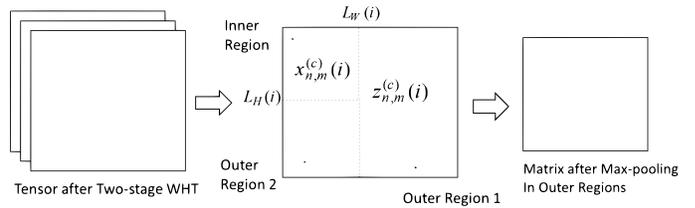}}
 \end{center}
 \vspace{-2mm}
 \caption{The proposed WHT-based region partition proposal and the regional max-pooling operation on the WHT transform-domain images. The original images with three RGB channels are firstly transformed by two-stage WHT, then are partitioned to the inner region and the outer region per image. The max-pooling operation is further performed in the partitioned regions.}\label{WHT_Region_Adaptation}
 \vspace{-3mm}
 \end{figure}

We present now our solutions to address the challenges of object classification in underwater environments, including the fixed-region and the adaptive-region cases.

\textbf{Fixed Region Proposal:}
In the proposed image feature compression, first, the underwater object image tensor data is passed through a column-wise Walsh-Hadamard Transformation~(WHT), afterwards, the transformed tensor data is passed through a row-wise WHT transformation. The two-stage WHT is utilized on the three GRB channels and the post-WHT RGB channel matrices of size $N$-by-$N$-by-$3$ are obtained. Then the max-pooling is performed on the post-WHT RGB channel matrices. The resulting tensor is partitioned into an inner region and two outer regions, namely outer region~1 and outer region~2. In the inner region, only max-pooling in three RGB channels is performed. In outer region~1, the max-pooling is done in matrices, while in outer region~2, the max-pooling is done in vectors. The proposed image feature compression method is depicted in Fig.~\ref{WHT_Region_Adaptation}. The procedure includes the steps of column-wise and row-wise WHT on the RGB channels in the original image, the max-pooling in RGB channels in both inner and outer regions, and the max-pooling in tensor matrices in outer regions. 

This proposed method is motivated by the fact that the elements in the inner region are much larger than the elements in the outer region. In other words, the elements in the inner region contain much richer information about the image than the elements in outer regions. In addition, \textit{due to the scattering and coloring effects in underwater, the obtained images are mostly of a single color. Thus, it is meaningful to compress three color channels into one channel in underwater image datasets.} Therefore, in the inner region, the RBG channels are max-pooled while all elements are preserved. In outer regions, the max-pooling operations are done on the tensor elements in the two partitioned regions. In the fixed-region proposal, the size of the inner region is a predefined parameter. This size is chosen by observing the amplitude values in the two-stage WHT transform domain vector. The amplitude peaks are included in the inner region, and the outer regions contain the low-amplitude and noisy values after the transformation.

\begin{figure*}[t!]
 \begin{center}
 \scalebox{0.56}
 {\includegraphics[height=6.5in, width=11in]{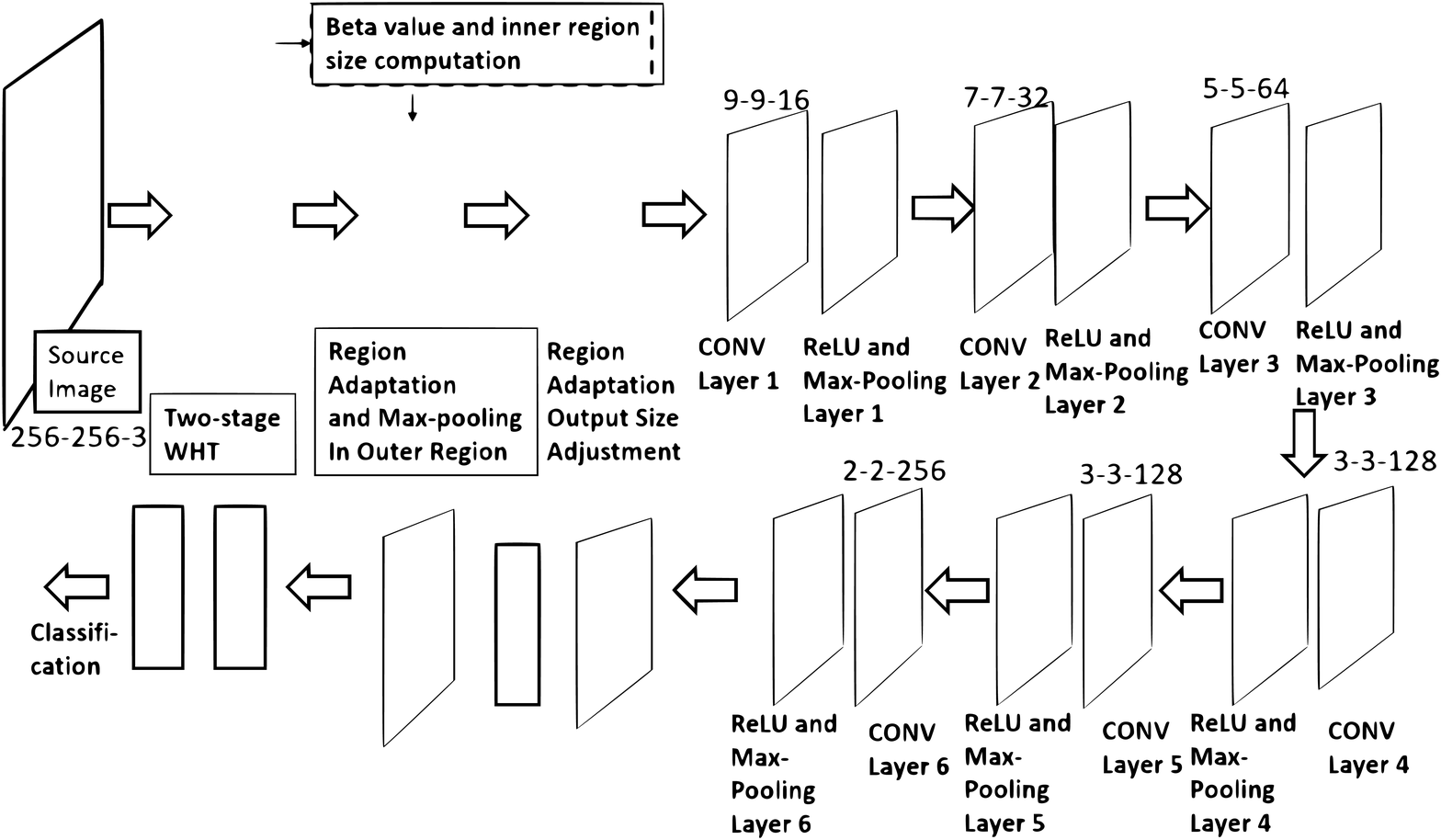}}
 \end{center}
 \vspace{-5mm}
 \caption{The CNN architecture with the adaptive region proposal is depicted. The beta values are computed based on the input image data, and the image inner region and outer regions are further determined based on the $beta(i)$ values. The max-pooling operations are performed on the outer region, and the output tensor size is further adjusted before CONV layer processing.}\label{CNN_Arch_Region_Adaptation}
 \vspace{-3mm}
 \end{figure*}

The proposed method is distinguished from the conventional WHT in the sense that, the resulting two-stage WHT is processed by the regional max-pooling to further reduce the dimensions of the resulting matrix; therefore, we speed up the processing of the CNN classification structure. Assume the input image matrix $\bf{X}$ has size $N_H$-by-$N_W$-by-$N_C$, where $N_H$ and $N_W$ are the height and width values, respectively, and $N_C$ is the number of color channels. The first stage transformation is a linear operation, i.e.,
\begin{equation}
    {\bf{U}} = {\bf{W}}_1 {\bf{X}},
\end{equation}
where ${\bf{W}}_1$ is the Walsh-Hadamard matrix of size $N_H$-by-$N_H$. The operation is done for the $N_C$ color channels producing the transform-domain matrix ${\bf{U}}$ with size $N_H$-by-$N_W$-by-$N_C$. The second transformation is done as,
\begin{equation}
    {\bf{V}} = {\bf{U} \bf{W}}_2,
\end{equation}
where the Walsh-Hadamard matrix ${\bf{W}}_2$ has size of $N_W$-by-$N_W$. The operation is performed for all the $N_C$ color channels. The resulting transform-domain matrix ${\bf{V}}$ has size $N_H$-by-$N_W$-by-$N_C$. The matrix is further written as,
\begin{equation}
    {\bf{V}} = \left[ {\begin{array}{*{20}c}
   {{\bf{R}}_I }  \\
   {{\bf{R}}_{O2} }  \\
\end{array}|{\bf{R}}_{O1} } \right],
\end{equation}
where ${{\bf{R}}_I }$ is the matrix of inner region and has size $L_H$-by-$L_W$-by-$N_C$. ${{\bf{R}}_{O1} }$ is the matrix of outer region 1 with size $N_H$-by-$(N_W-L_W)$-by-$N_C$. ${{\bf{R}}_{O2}}$ is the matrix of outer region 2 with size $(N_H-L_H)$-by-$L_W$-by-$N_C$. The values of $L_H$ and $L_W$ are fixed values for the fixed region proposal, and are varied values for the adaptive region proposal.

The following operations are performed in two regions. 
\begin{enumerate}

\item In the inner region, the max-pooling operation is performed on three RGB channels only. Denote the max-pooling operation on the $N_C$ color channels in the inner region by $g_0()$, the inner region matrix after max-pooling is ${{\bf{\hat R}}_I } = g_0({{\bf{R}}_I })$, resulting the matrix of size $L_H$-by-$L_W$. 

\item In the outer regions, the max-pooling on both the RGB channels and the data matrices are performed. For the outer region ${{\bf{R}}_{O1} }$, the max-pooling is performed with size $N_M$-by-$N_M$ square, while for ${{\bf{R}}_{O2} }$, it is done with size $N_V$-by-1 vector. Here, the parameters $N_M$ and $N_V$ are the chosen max-pooling parameters. Define $g_1()$ as the max-pooling operation on tensor ${{\bf{R}}_{O1} }$, and $g_2()$ as max-pooling operation on tensor ${{\bf{R}}_{O2} }$. The matrices after the max-pooling operations are denoted by ${{\bf{\hat R}}_{O1} } = g_1 ( {{\bf{R}}_{O1} })$, and ${{\bf{\hat R}}_{O2} } = g_2 ( {{\bf{R}}_{O2} })$. 
\end{enumerate}

The resulting matrix is the concatenated matrix by the matrices ${{\bf{\hat R}}_{O1} }$, ${{\bf{\hat R}}_{O2} }$ and ${{\bf{\hat R}}_I }$ with zero-padding in the null elements to form the rectangular-shaped output transform-domain matrix. Note that, in the following adaptive region proposal, the sizes of the matrices ${{\bf{\hat R}}_{O1} }$ and ${{\bf{\hat R}}_{O2} }$ are determined by the algorithm of the adaptive region proposal.

\textbf{Adaptive Region Proposal:}
In the adaptive region method, the input images are firstly transformed by a two-stage WHT transformation process. For each output tensor of the two-stage WHT with index $i$, the matrix is partitioned into inner region and outer region. The width $L_W (i)$ and height $L_H (i)$ of the inner regions are adaptively determined per input image matrix. Define the post-WHT elements in the inner region by by $x_{n,m}^{(c)} (i)$ on the $n$-th row, $m$-th column in the $c$-th color channel of the $i$-th image. Further define the post-WHT elements in the outer region by $z_{n,m}^{(c)} (i)$ on the $n$-th row and $m$-th column on the $c$-th color channel of the $i$-th image. Introduce a metric $\beta(i)$ defined by,
\begin{equation}
   \beta (i) = \frac{{\sum\limits_{n,m,c}^{} {\left| {x_{n,m}^{(c)} (i)} \right|^2 } }}{{\sum\limits_{n,m,c}^{} {\left| {x_{n,m}^{(c)} (i)} \right|^2 }  + \sum\limits_{n,m,c}^{} {\left| {z_{n,m}^{(c)} (i)} \right|^2 } }}.
\end{equation}

\begin{figure*}[t!]
        \centering
        \begin{subfigure}
            \centering
            \includegraphics[height=1.2in]{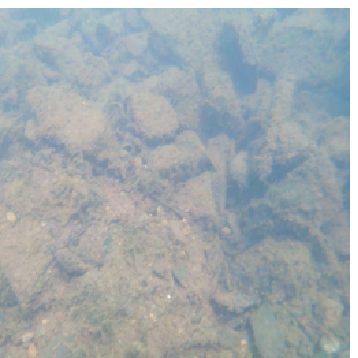}
        \end{subfigure}%
~
        \begin{subfigure}
            \centering
            \includegraphics[height=1.2in]{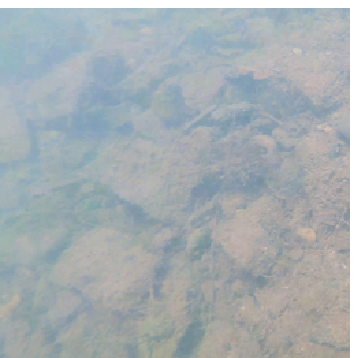}
        \end{subfigure}
        ~
        \begin{subfigure}
            \centering
            \includegraphics[height=1.2in]{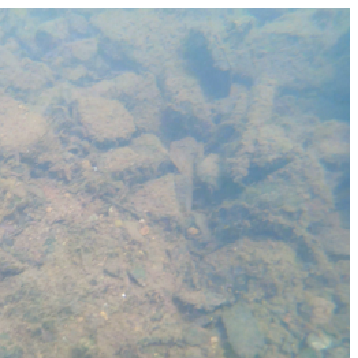}
        \end{subfigure}
        ~ 
        \begin{subfigure}
            \centering
            \includegraphics[height=1.2in]{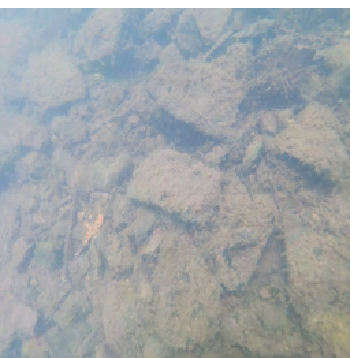}
        \end{subfigure}
        ~
        \begin{subfigure}
            \centering
            \includegraphics[height=1.2in]{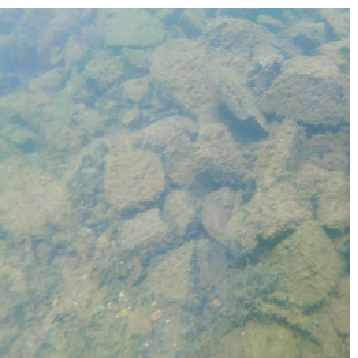}
        \end{subfigure}
        ~ \\
        \begin{subfigure}
            \centering
            \includegraphics[height=1.2in]{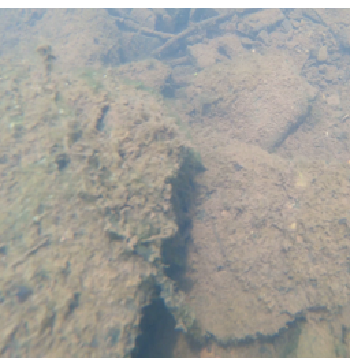}
        \end{subfigure}
                \begin{subfigure}
            \centering
            \includegraphics[height=1.2in]{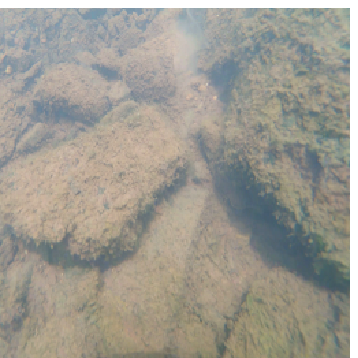}
        \end{subfigure}
        ~
        \begin{subfigure}
            \centering
            \includegraphics[height=1.2in]{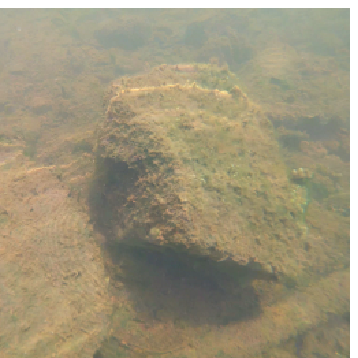}
        \end{subfigure}
        ~ 
        \begin{subfigure}
            \centering
            \includegraphics[height=1.2in]{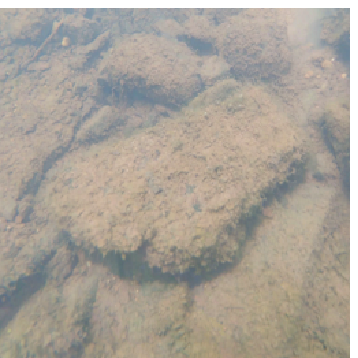}
        \end{subfigure}
        ~
        \begin{subfigure}
            \centering
            \includegraphics[height=1.2in]{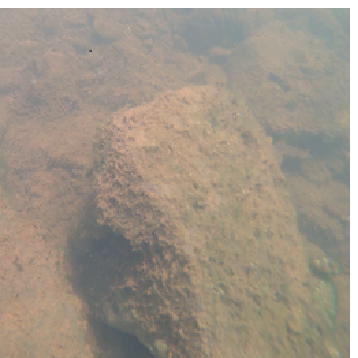}
        \end{subfigure}
        ~ \\
        \caption{\label{fig:rock_images} Sample rock images: the first row demonstrates the small rocks, while the second row represents large rocks.}
     \vspace{-0.15in}
\end{figure*}


\begin{figure*}[t!]
        \centering
        \begin{subfigure}
            \centering
            \includegraphics[height=1.2in]{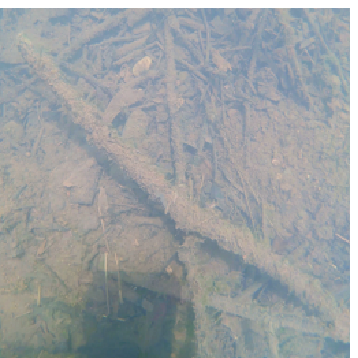}
        \end{subfigure}%
~
        \begin{subfigure}
            \centering
            \includegraphics[height=1.2in]{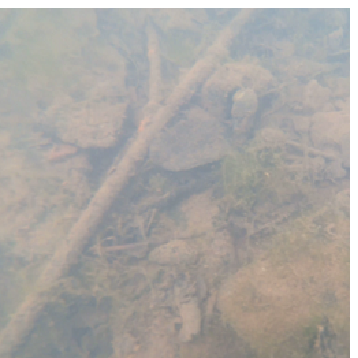}
        \end{subfigure}
        ~
        \begin{subfigure}
            \centering
            \includegraphics[height=1.2in]{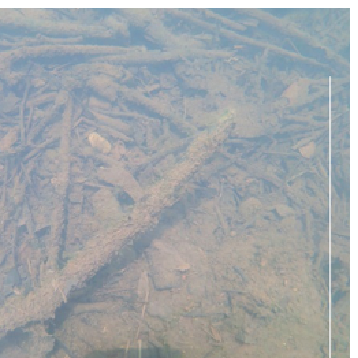}
        \end{subfigure}
        ~ 
        \begin{subfigure}
            \centering
            \includegraphics[height=1.2in]{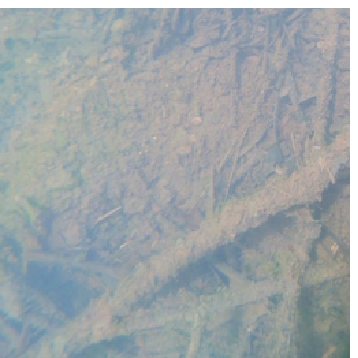}
        \end{subfigure}
        ~
        \begin{subfigure}
            \centering
            \includegraphics[height=1.2in]{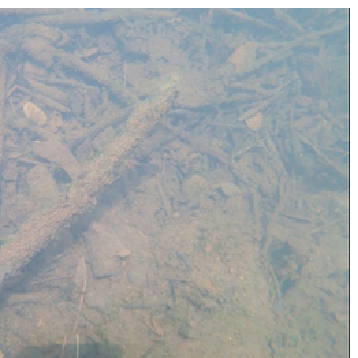}
        \end{subfigure}
        ~ \\
        \begin{subfigure}
            \centering
            \includegraphics[height=1.2in]{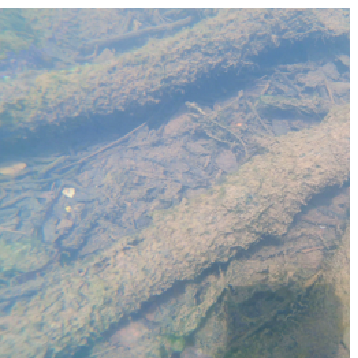}
        \end{subfigure}
                \begin{subfigure}
            \centering
            \includegraphics[height=1.2in]{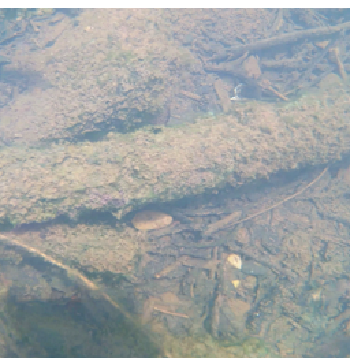}
        \end{subfigure}
        ~
        \begin{subfigure}
            \centering
            \includegraphics[height=1.2in]{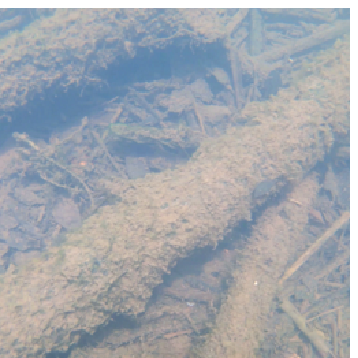}
        \end{subfigure}
        ~ 
        \begin{subfigure}
            \centering
            \includegraphics[height=1.2in]{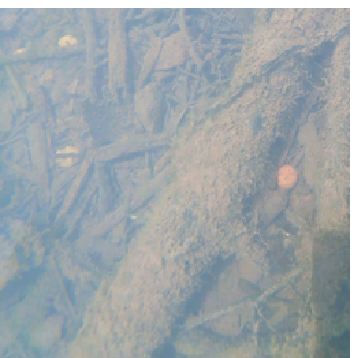}
        \end{subfigure}
        ~
        \begin{subfigure}
            \centering
            \includegraphics[height=1.2in]{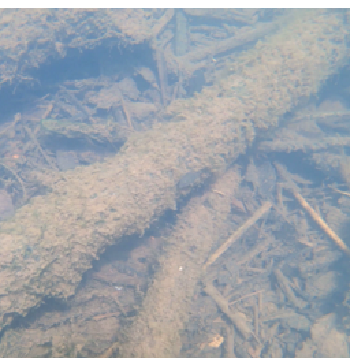}
        \end{subfigure}
        ~ \\
        \caption{\label{fig:wood_images}Sample wood branch images: the first row images belong to small wood branches, while the second row show some large wood branches.}
     \vspace{-0.15in}
\end{figure*}

Define a threshold value of $\eta$. The parameters of $L_W (i)$ and $L_H (i)$ are determined by comparing the value $\beta(i)$ with this threshold value $\eta$. The threshold should be chosen based on the training data and should maximize the performance gain. The performance metrics are the converged training accuracy and the validation accuracy. It is expected that with adaptive region, the converged training accuracy and the validation accuracy results can be further improved compared with the approach of fixed inner and outer region sizes. 

The choice of the threshold value $\eta$ is different based on the type of objects in the images. The object images with high level of details including images of small rocks, require larger value of threshold $\eta$, compared with the object images with low level of details including images of large rocks. The images of small wood branches require larger threshold $\eta$ compared with images of large wood branches. This observation is related to the amplitude distribution of the elements in the transform-domain matrix. $\eta$ is chosen manually at the testing phase of the program. 

The CNN architecture with the adaptive region proposal is depicted in Fig.~\ref{CNN_Arch_Region_Adaptation}. The adaptive region is designed before the first CONV layer to adapt the sizes of inner regions per input image. The input image data is processed to compute the $\beta$ values and to compare with the threshold value $\eta$ to determine the sizes of the inner and outer regions. The regional max-pooling operation is performed in the outer regions in the resulting matrices. The beta values and inner region sizes are computed based on the input image data and the adaptive region algorithm. Due to the constraint of the CONV layers implemented on Keras with Tensorflow backend, the input sizes of images by the adaptive region need to be adjusted so that the CONV layers can process. The zero-padding operation is performed on the resulting matrices. This CNN architecture is a basic architecture to verify the performance advantage of the proposed work. Note that, the depth, number of layers and hyperparameters of each layer can be adjusted according to the application requirement. This CNN architecture is chosen to fit our underwater image classification application. 

There are several source compression methods with optimization-based algorithm designs. The Gaussian mixture models (GMM) with the Lloyd clustering algorithm is studied for the image source compression~\cite{Aiyer05}. The dual augmented Lagrangian method~(DALM) with Principal Component Analysis~(PCA) pre-filtering is proposed~\cite{Davis19} to improve the object classifications accuracy for noisy images. The image compression techniques are proposed by the hash kernel based dimensionality reduction and product quantizer based encoding~\cite{Sanchez11} for large-scale image classification. These existing approaches require an optimization framework to perform the image source compression. In contrast, our proposal only requires linear matrix multiplication. Therefore the computational complexities of these existing approaches are significantly higher than our approach. These approaches have different assumptions in designing their algorithm. The source compression algorithms in these existing works have iterative procedures in the data compression, therefore have significant higher level of computational complexity. In contrast, our proposal is based on linear transformation with a simple non-linear region partitioning. The proposal has much lower computational complexity compared with the existing approaches.

\section{Performance Evaluation}\label{performance}
We describe the experiment setup and present the results.

\textbf{Experiment Setup:}
We performed multiple rounds of experiments in the Raritan River-New Jersey during Summer'19. We modified the BlueROV2, a remotely operated underwater vehicle developed by BlueRobotics~Inc., for implementing the scenario. This vehicle is equipped with four vertical vectored thrusters, enabling the robot to reach a maximum speed of $1~\rm{m/s}$. This underwater robot is also equipped with a 1080p camera with a tilt range of $90$ degree. Sample images of the four types of objects are depicted in Fig.~\ref{fig:rock_images} for small and large rocks, and in Fig.~\ref{fig:wood_images} for small and large wood branches.

\textbf{Results:}
The proposal is evaluated with the CNN structure with the adaptive region method at the front layers of this architecture. There are 6 subsequent convolutional layers with ReLU activation and max-pooling. The structure at the output of the neural network is composed of dense, flatten, dropout, and ReLU activation layers. The proposal is evaluated on TensorFlow/Keras software package. The TensorFlow/Keras library supports Stochastic Gradient Descent~(SGD) based training of CNN. In our application of underwater object classification, the objects of underwater rocks and underwater wood branches are not the daily objects in any pre-trained CNN models. Therefore, the application-specific CNN model needs to be re-trained for this underwater object classification. Because we are proposing the source compression method, the comparison done in this work is to compare the proposed source compression with existing source compression approaches. This comparison is performed on the same CNN architecture. The CNN architecture parameters including the number of convolutional layers and layer filter sizes are tuned to fit the underwater object dataset. The standard CNN architectures are not adopted in our evaluation, because the structural parameters of these standard architectures are optimized based on general public datasets, not based on our underwater object datasets.

The parameter $\lambda$ is defined to be the number of image samples per epoch and the parameter $\gamma$ is defined to be the batch size, i.e., the number of training samples in processing before the SGD updating. These parameters can affect the computation time. The larger the $\lambda$ value, the longer the computation time; in contrast, the larger the $\gamma$ value, the shorter the computation time. In the computer experiments, the parameter $\lambda$ is chosen for values of $\lambda = 100$, $\lambda = 200$, and larger values of $\lambda = 500$. The parameter $\gamma$ is chosen to be $8$, $16$, and $32$. The parameter $\lambda$ is chosen from $100$ to $500$. The number of epochs is $100$. The parameter $\gamma$ is chosen from $8$, $16$, $32$, $64$ or $96$. For CONV layer 1, there are 16 filers of size 9-by-9; For CONV layer 2, there are 32 filers of size 7-by-7; for CONV layer 3 and 4, there are 64 filters of size 5-by-5;  for CONV layer 5 and 6, there are 128 filters of size 3-by-3. For all the CONV layers, the activation is ReLU, and the max-pooling layer pool size is 2-by-2 square.

\begin{figure*}[t!]
        \centering
        \hspace{-5mm}
        \begin{subfigure}
            \centering
            \includegraphics[height=1.8in]{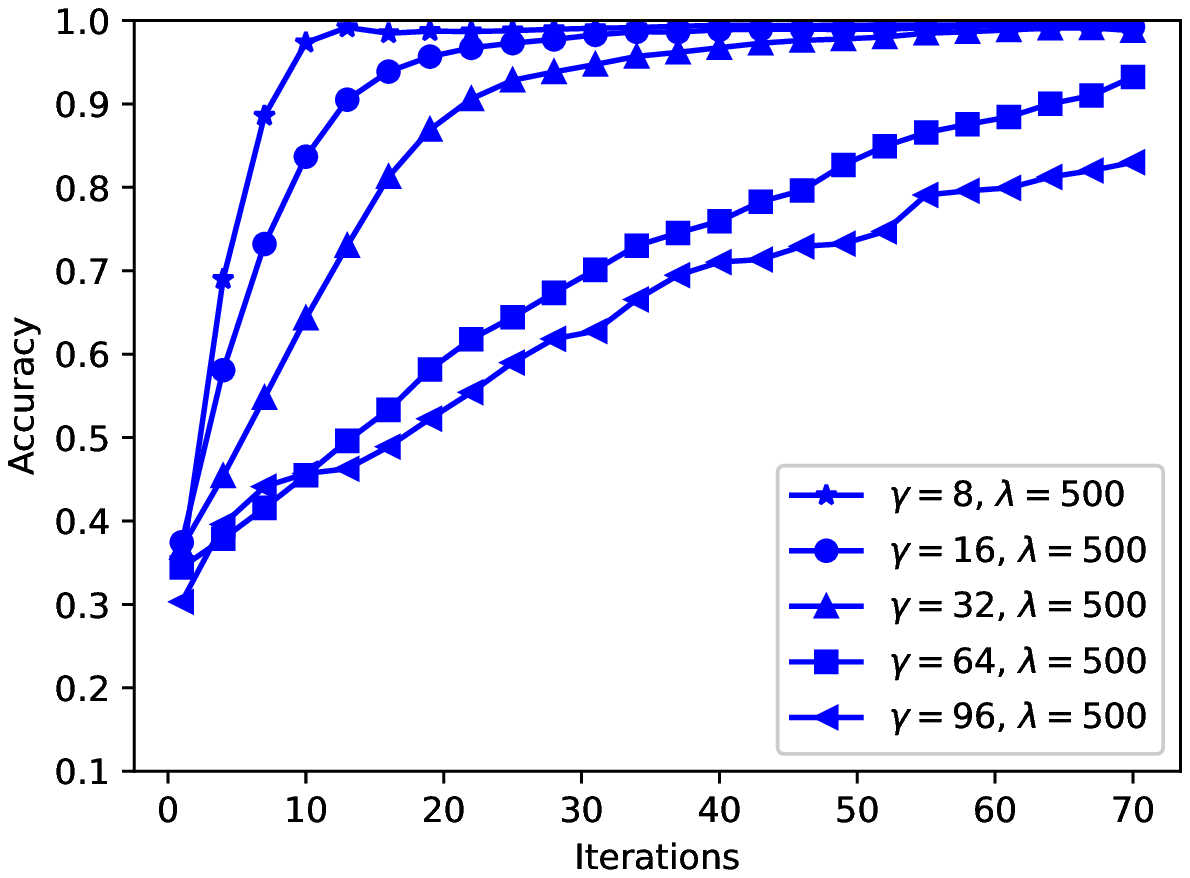}
            \label{fig:FixedRegion_Vary_Gamma_Opt}
            \vspace{3mm}
        \end{subfigure}
~ \hspace{-10mm}
        \begin{subfigure}
            \centering
            \includegraphics[height=1.8in]{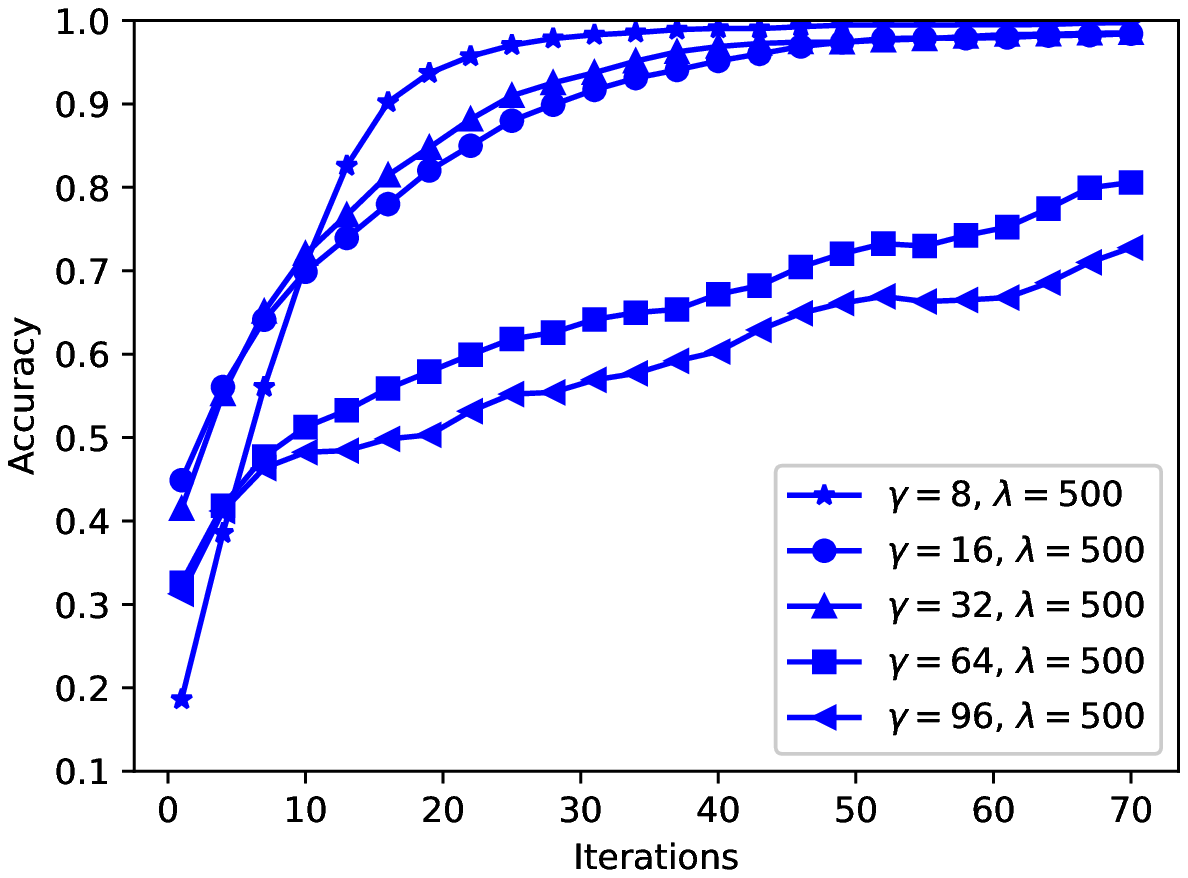}
            \label{fig:WHT_Vary_Gamma_Opt}
        \end{subfigure}
~ \hspace{-11mm}
        \begin{subfigure}
            \centering
            \includegraphics[height=1.8in]{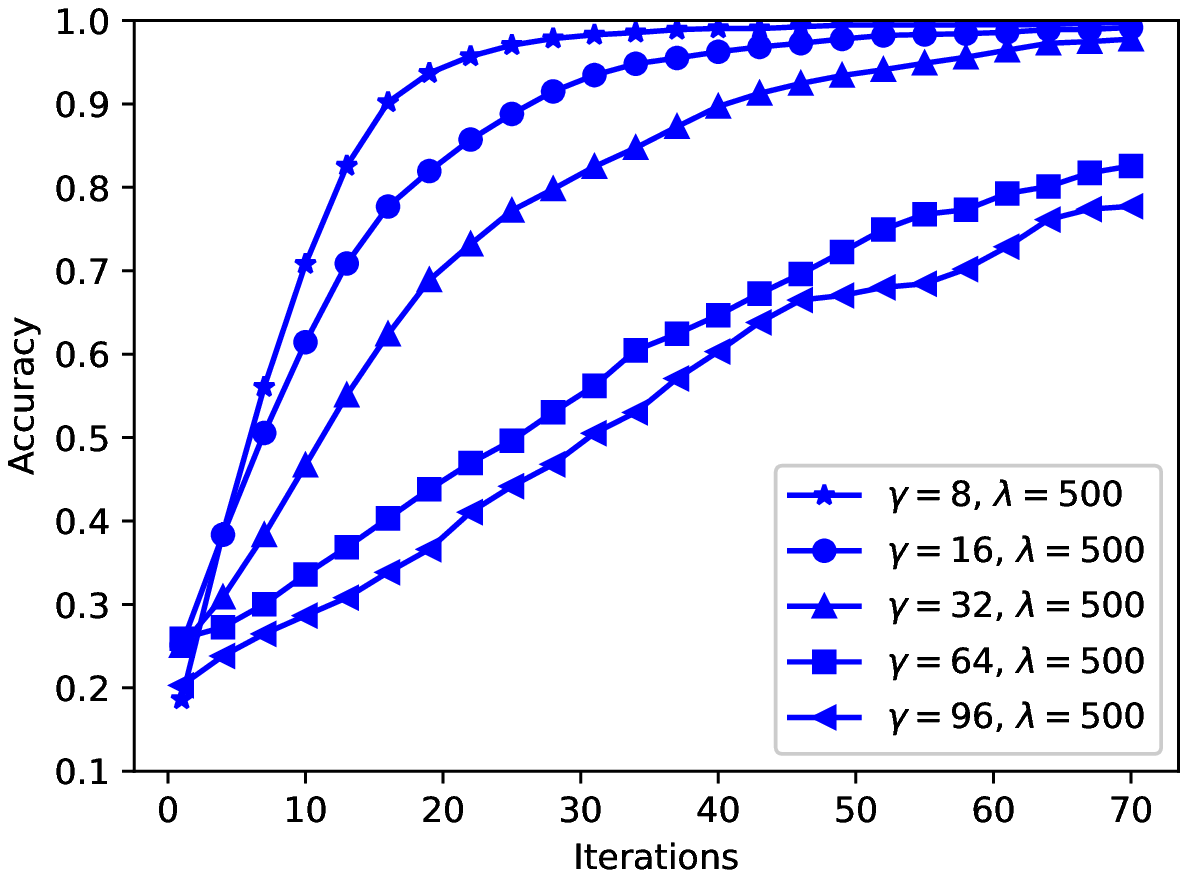}
            \label{fig:DCT_Vary_Gamma_Opt}
        \end{subfigure}
       \hspace{0.3cm} (a)   \hspace{5cm} (b)     \hspace{5cm} (c)
        \caption{\label{fig:conv_varied_gamma} The validation accuracy convergence curve of (a)~The fixed region proposal; (b)~The WHT method; (c)~The DCT method. Parameter $\lambda$ is $500$, and the parameter $\gamma$ is varied from $8$ to $96$.}
\end{figure*}

\begin{figure*}[t!]
        \centering
        \hspace{-5mm}
        \begin{subfigure}
            \centering
            \includegraphics[height=1.8in]{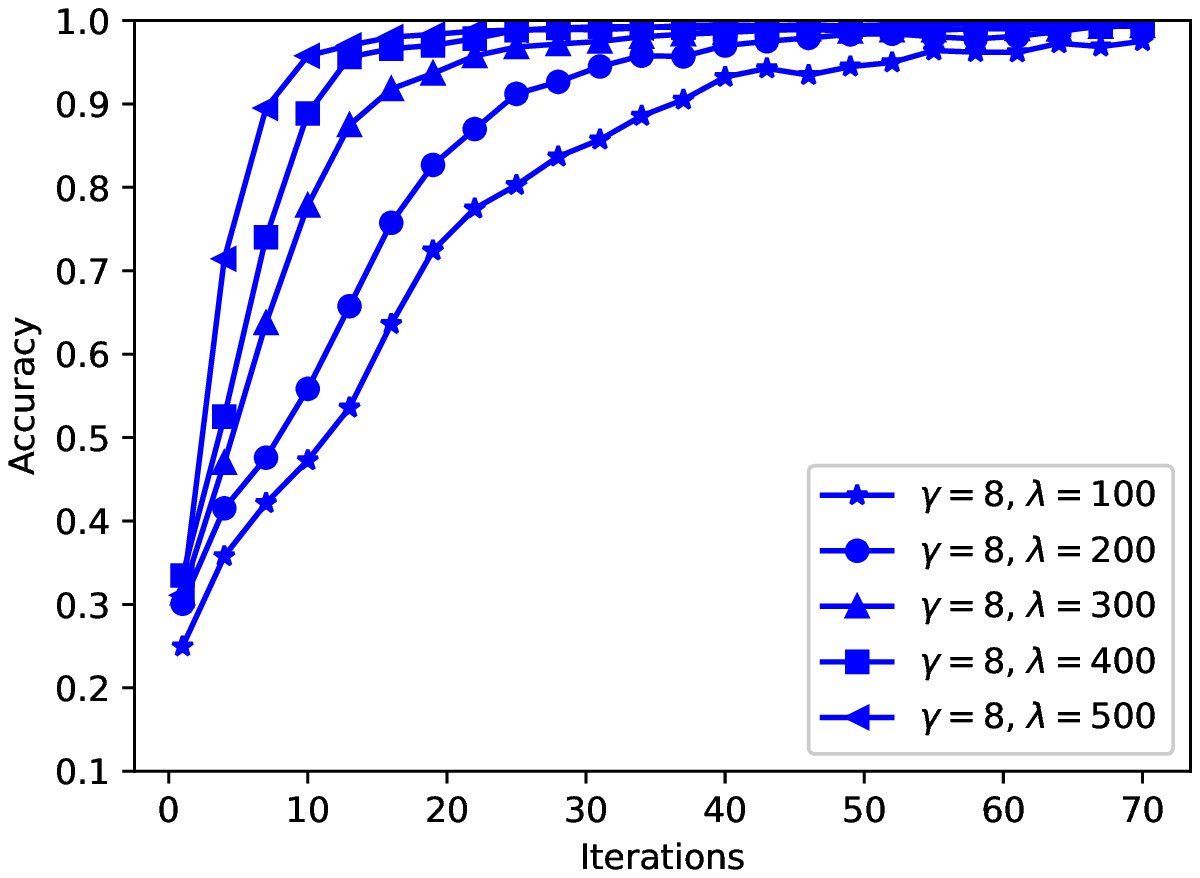}
            \label{fig:FixedRegion_Vary_Lambda_Opt}
            \vspace{3mm}
        \end{subfigure}
~ \hspace{-10mm}
        \begin{subfigure}
            \centering
            \includegraphics[height=1.8in]{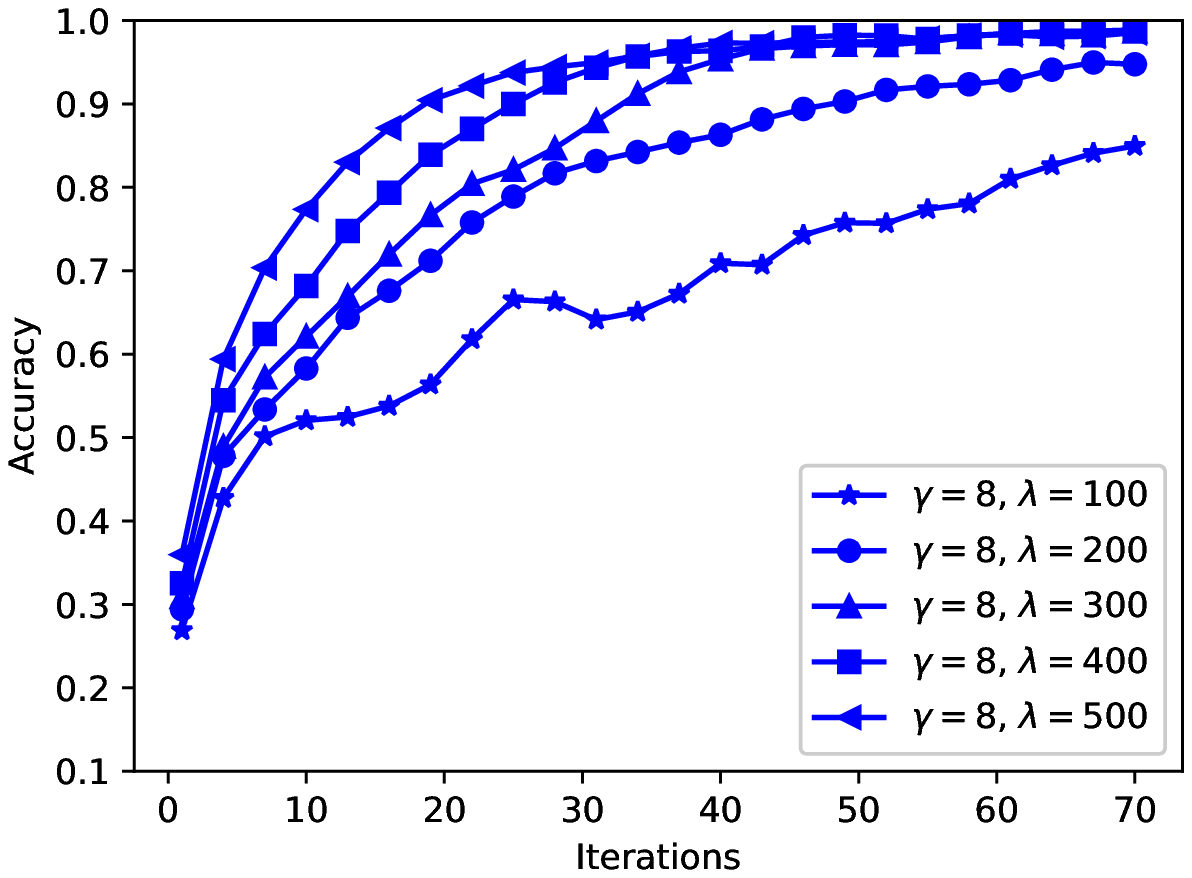}
            \label{fig:WHT_Vary_Lambda_Opt}
        \end{subfigure}
~ \hspace{-11mm}
        \begin{subfigure}
            \centering
            \includegraphics[height=1.8in]{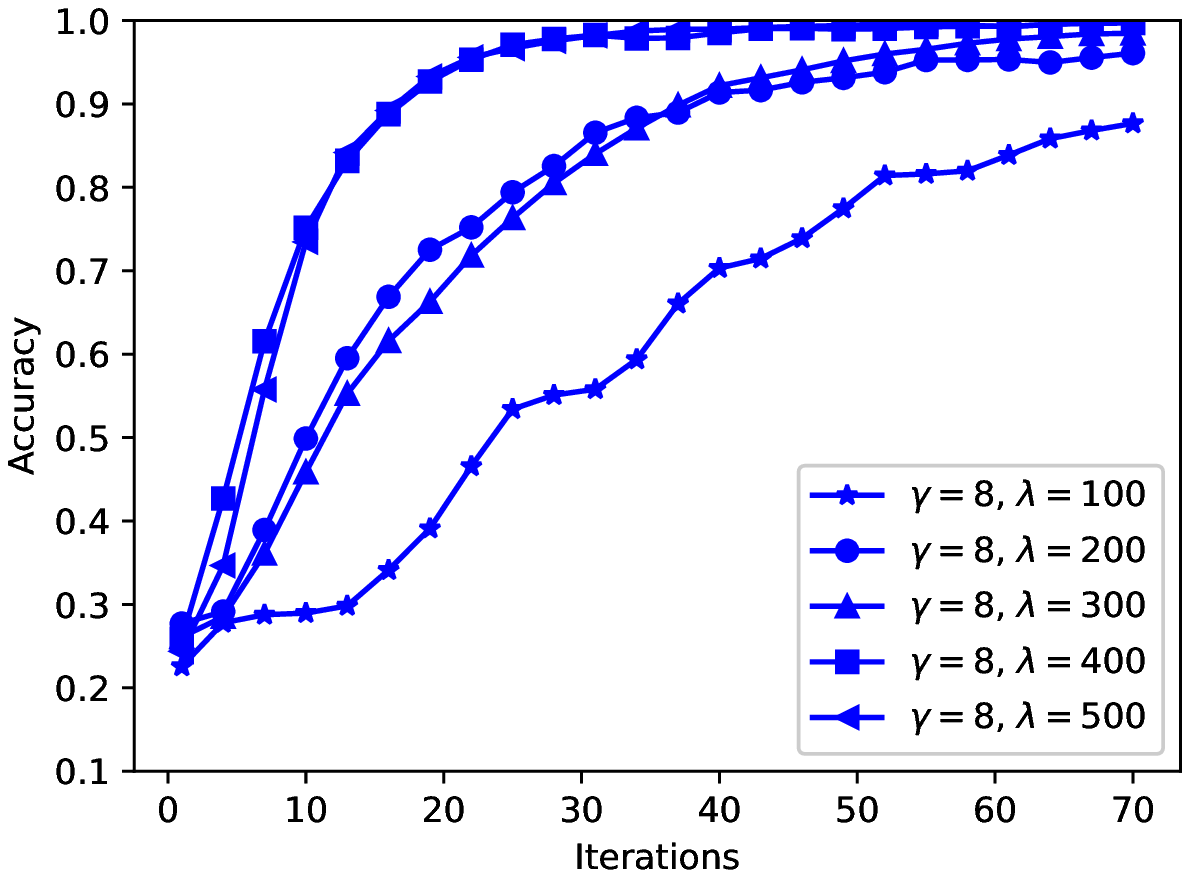}
            \label{fig:DCT_Vary_Lambda_Opt}
        \end{subfigure}
        \hspace{0.1cm} (a)   \hspace{5cm} (b)     \hspace{5cm} (c)
        \caption{\label{fig:convergence_lambda} The validation accuracy convergence curve of (a)~The fixed region proposal; (b)~The WHT method; (c)~The DCT method. Parameter $\gamma$ is $16$ and $\lambda$ varies from $100$ to $500$.}
\end{figure*}

The input images has parameters of image width $N_W = 256$, image height $N_H = 256$ and the number of color channels $N_C = 3$. The proposal is performed by the following procedure. The 256-by-256-by-3 image matrices are transformed by two-stage WHT, resulting in 256-by-256-by-3 transformation-domain matrices. There are two approaches to partition the regions: 

$(1)$ For the proposal of fixed inner region size, the inner region size is defined to be 32-by-32. The outer region is partitioned into two regions, outer region 1 and outer region 2. All the inner and outer regions are processed with max-pooling on three RGB channels, resulting in 256-by-256 matrix. The regional max-pooling parameters are chosen by $N_M = 4$, and $N_V=8$. The max-pooling in outer region 1 is performed by 4-by-4 square, resulting 56-by-88 matrix. The max-pooling in outer region 2 is done by 8-by-1 vector, resulting in 28-by-32 matrix. Zero-padding is utilized for the null areas in the final matrix. The final matrix is of size 64-by-88. With this proposed procedure, the 256-by-256-by-3 image matrices are converted to the 64-by-88 transform-domain matrices. 

$(2)$ For the adaptive region approach, $\beta$ values are computed and compared with the threshold value $\eta$. For a series of $N_{in}$ input images, define the maximum width after the max pooling of the fixed region proposal to be $N_{W,max} = \mathop {\max } \{ N^{(i)}_{W,MP} \}$ where $N^{(i)}_{W,MP}$ is the width after max-pooling for the $i$-th image, $i = 1,2,...,N_{in}$. The image data width values are padded with zeros to the size of $N_{W,max}$. The operations on the image height values are done by padding zeros to the size of $N_{H,max} = \mathop {\max } \{ N^{(i)}_{H,MP} \}$ where $N^{(i)}_{H,MP}$ is the height after max-pooling on the $i$-th image. This padding solution is to address the coding problem of variable image sizes. All the image matrices after adaptive region are padded to the same size to be processed by the CNN architecture. The resulting matrices are scaled by a constant value to adjust the amplitude of the data before sending to the CNN architecture. This constant is chosen to be $16$. The CONV layers and the parameters (filter sizes and the number of filters) have a direct impact on the classification accuracy convergence results for the adaptive region proposal. The CNN architecture under evaluation in this work has the proposed source feature compression method before the CONV layers to pre-process the data. The CNN architecture parameters are manually tuned to improve the performance. The proposal is compared with the WHT-only method and the DCT method under the same CNN architecture.

\begin{figure*}[t!]
        \centering
        \hspace{-5mm}
        \begin{subfigure}
            \centering
            \includegraphics[height=2.7in]{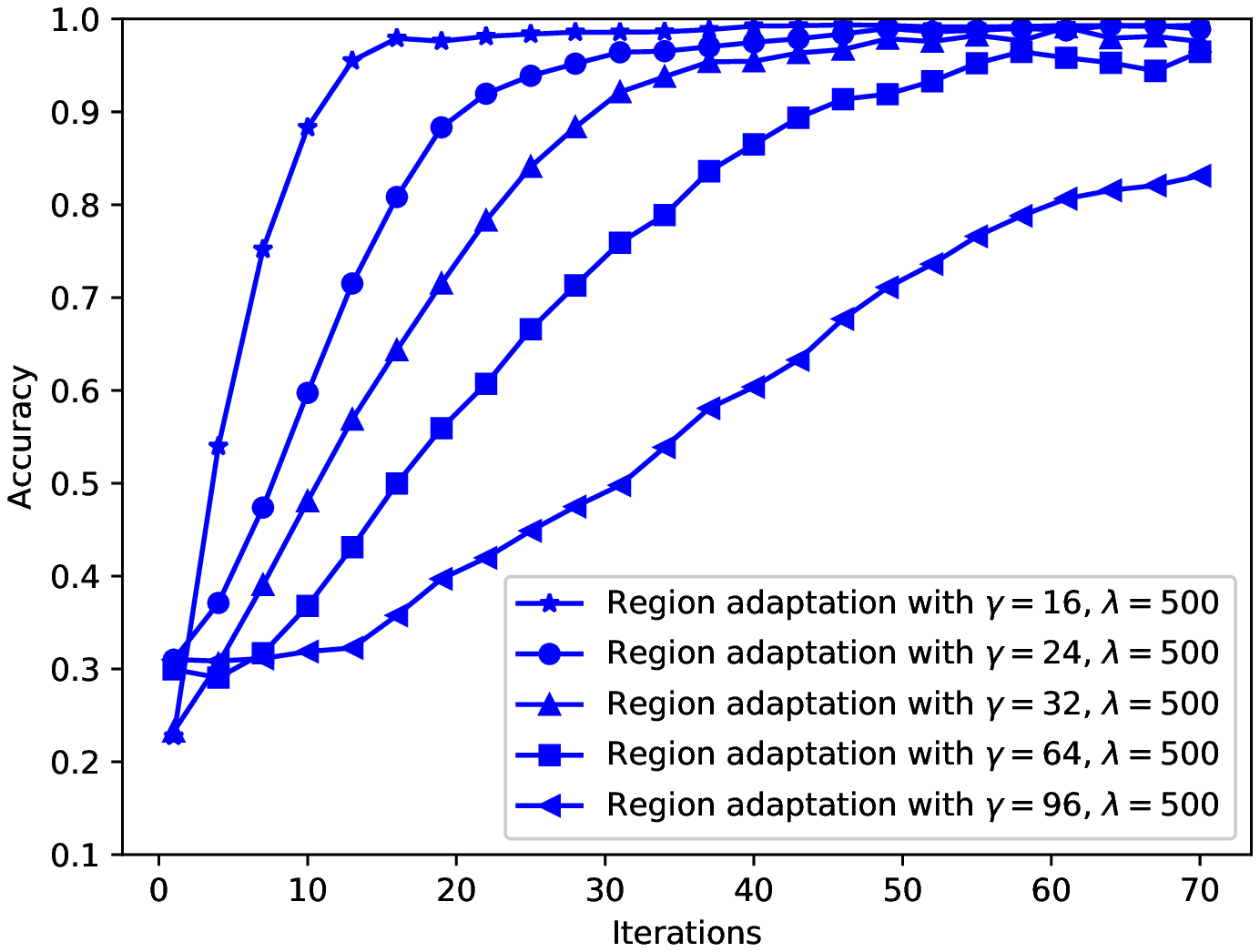}
            \label{fig:RegionGamma}
            \vspace{3mm}
        \end{subfigure}
~ \hspace{-10mm}
        \begin{subfigure}
            \centering
            \includegraphics[height=2.7in]{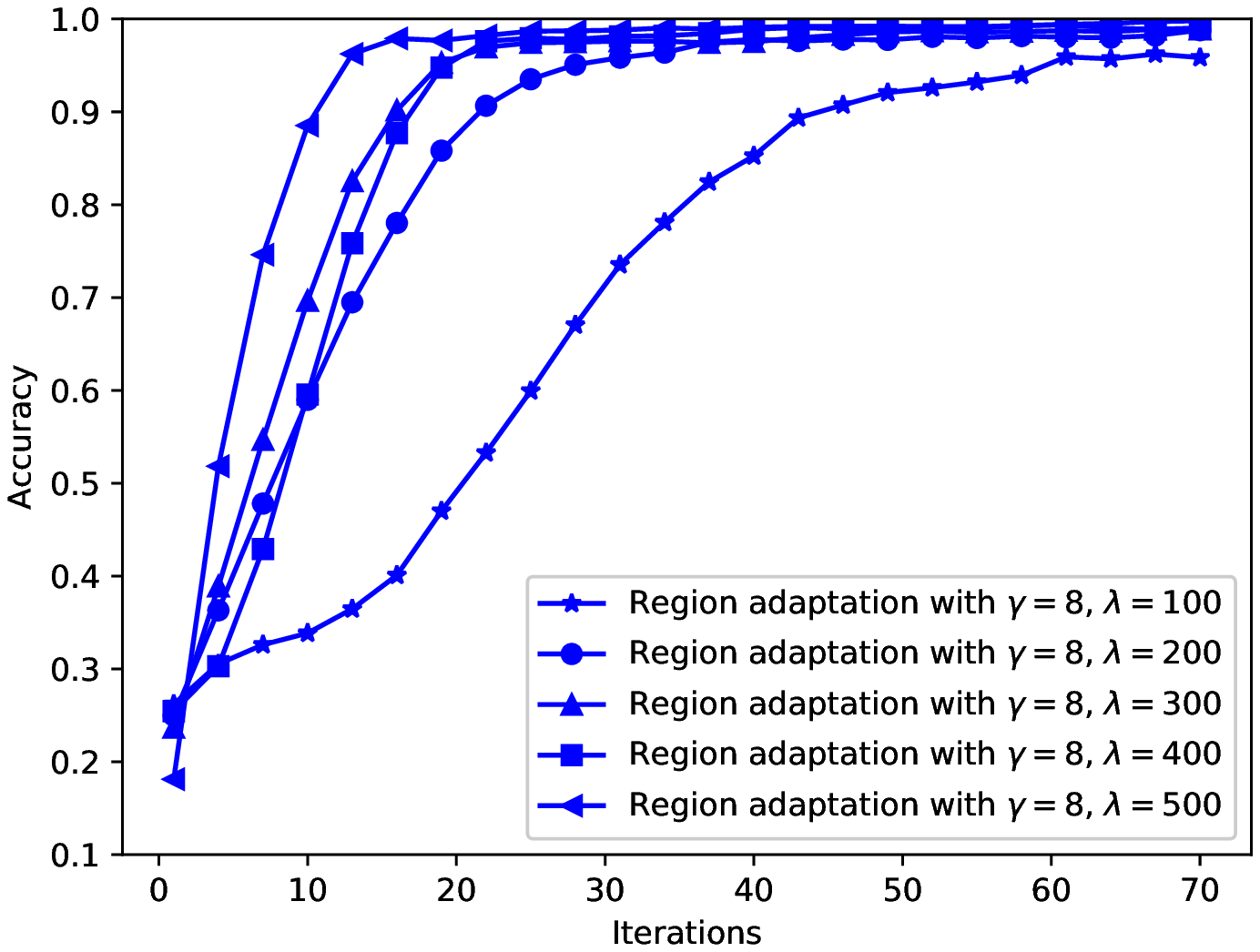}
            \label{fig:RegionLambda}
        \end{subfigure}
        \vspace{-2mm}
        \hspace{0.1cm} (a)   \hspace{7cm} (b)  
        \caption{\label{fig:Region_Adaptation} The classification accuracy convergence curve for (a)~varied $\gamma$ in the adaptive region proposal; (b)~varied $\lambda$ in the adaptive region proposal.}
\end{figure*}

Four types of objects are defined as classes: small rocks, large rocks, small wood branches and large wood branches. In the proposed method, there are source compression layers and the CNN classification layers. Our proposed fixed region proposal on the two-stage WHT tensor is at the source compression layers and targets to reduce the training computational complexity at the CNN classification layers. The training is performed with Stochastic Gradient Descent~(SGD) based iterative algorithm per epoch to optimize all the trainable parameters in the neural network. The proposal can reduce the training time due to the reduced data size; in the meanwhile, the regional max-pooling done at the outer regions has the effect of reducing the noise in the original images.

The training accuracy convergence curves of the proposal with fixed region size, the WHT-only method and the DCT method in comparison are given in Fig.~\ref{fig:conv_varied_gamma} for $\lambda = 500$ with the parameter $\gamma$ varying from $8$ to $96$, and Fig.~\ref{fig:convergence_lambda} for $\gamma = 8$ with parameter $\lambda$ varying from $100$ to $500$. From these figures, it can be observed that the proposed method has faster convergence concerning the iterations compared with the WHT-only method and the DCT method. This convergence speed improvement is due to the regional max-pooling operation introduced at the design of the proposed source compression method. With increased $\lambda$ values, the convergence curves of all methods are improved. The proposal has the best convergence among the three methods for the same $\lambda$ value due to the design of the fixed region proposal operation.

\begin{figure*}[t!]
        \centering
        \hspace{-5mm}
        \begin{subfigure}
            \centering
            \includegraphics[height=1.8in]{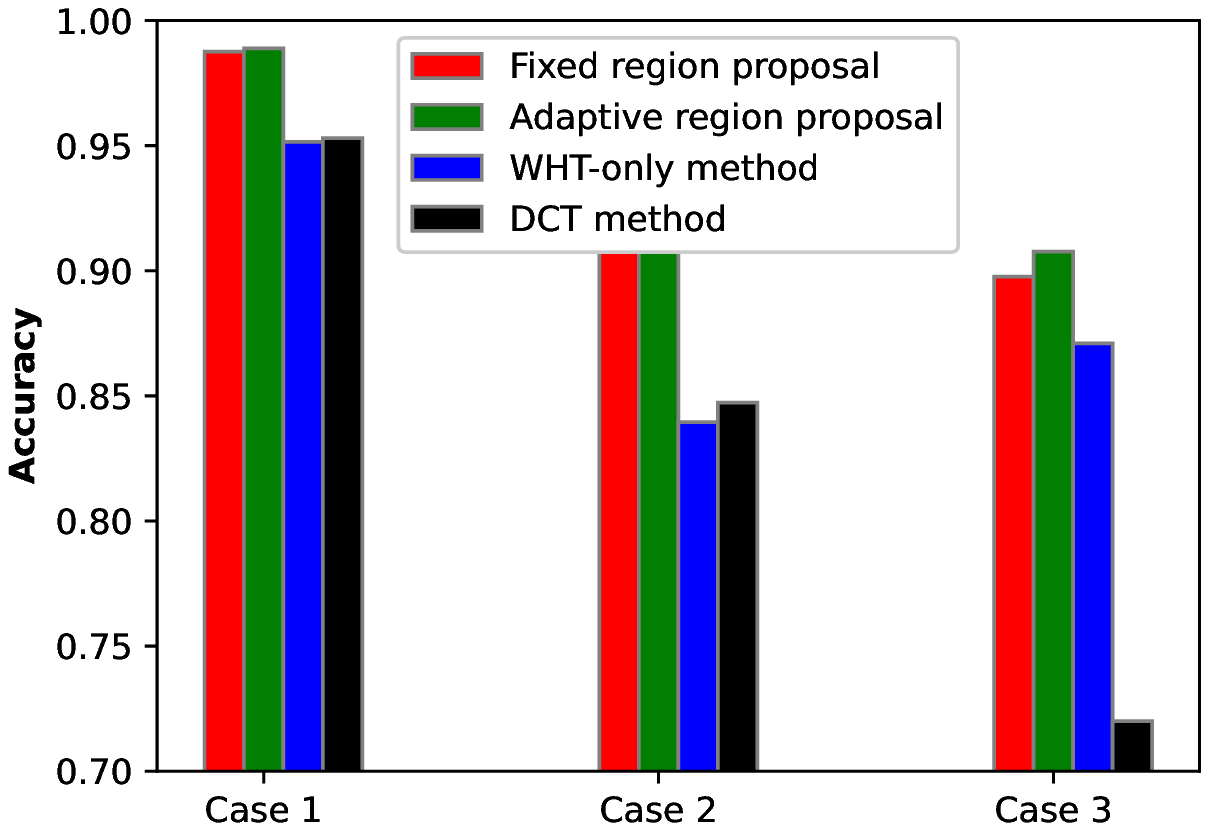}
            \label{fig:Comparison_FourMethods_Iter20}
            \vspace{3mm}
        \end{subfigure}
~ \hspace{-10mm}
        \begin{subfigure}
            \centering
            \includegraphics[height=1.8in]{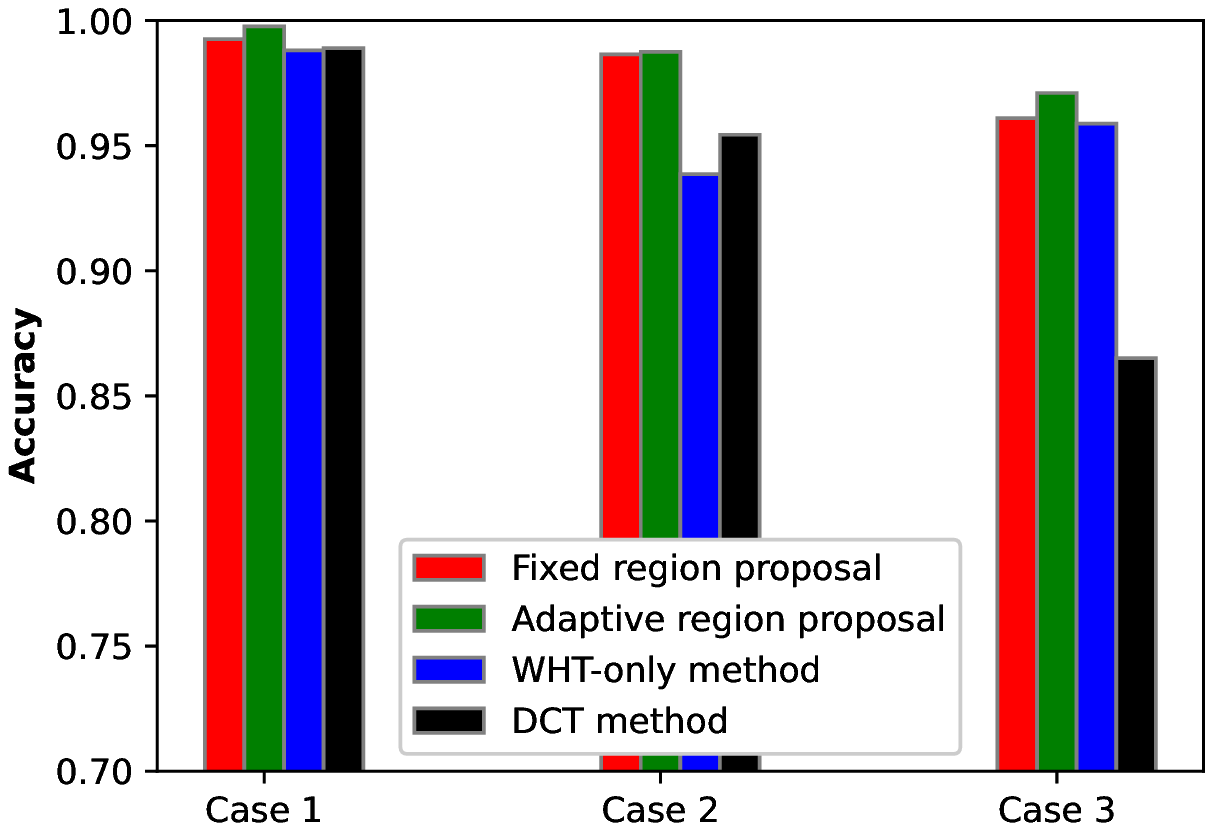}
            \label{fig:Comparison_FourMethods_Iter35}
        \end{subfigure}
~ \hspace{-11mm}
        \begin{subfigure}
            \centering
            \includegraphics[height=1.8in]{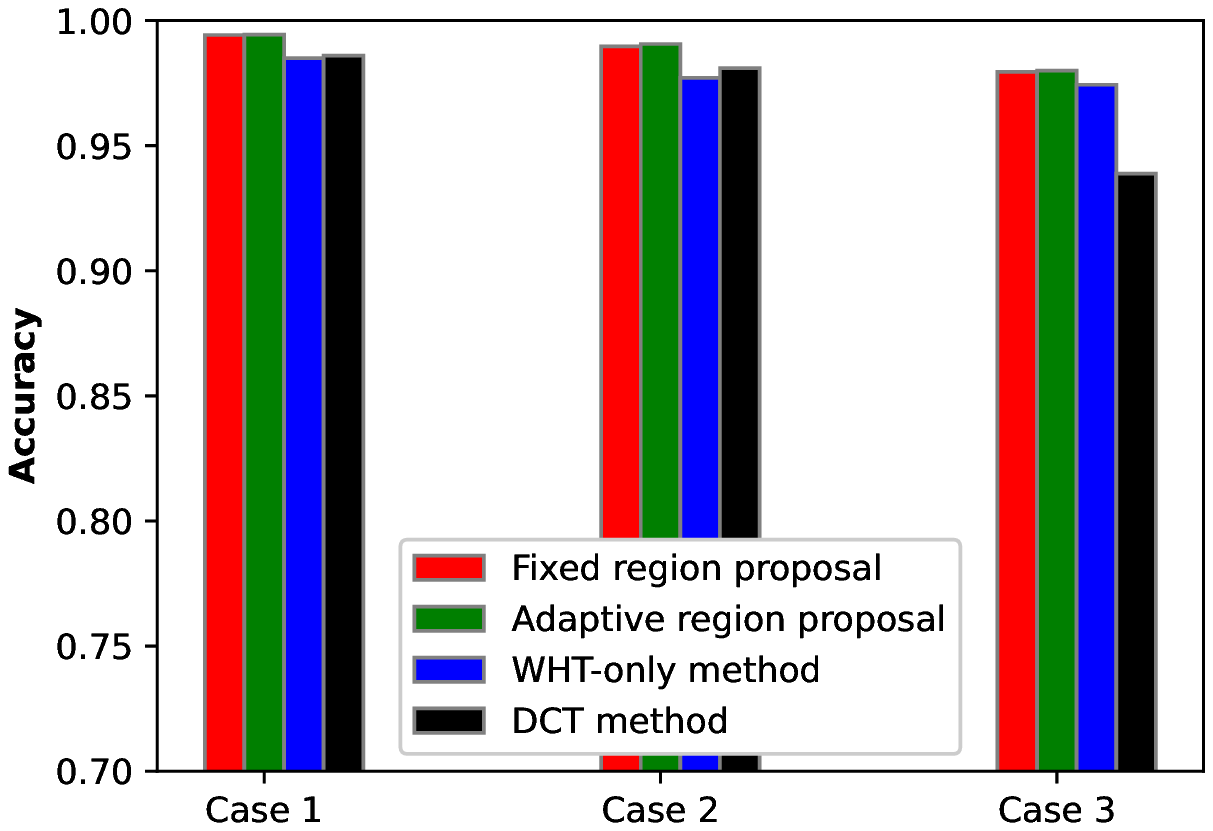}
            \label{fig:Comparison_FourMethods_Iter50}
        \end{subfigure}
        \hspace{0.2cm} (a)   \hspace{5cm} (b)     \hspace{5cm} (c)
        \caption{\label{fig:comparison_1} The comparison of four methods including fixed region proposal, adaptive region proposal, WHT-only method and DCT method. The three cases include case 1 of $\gamma = 8$, case 2 of $\gamma = 16$, and case 3 of $\gamma = 32$, with parameter (a)~iterations = 20; (b)~iterations = 35; (c)~iterations = 50.}
\end{figure*}

The adaptive region proposal is further evaluated and the results are depicted in Fig.~\ref{fig:Region_Adaptation}, including the convergence curves of the adaptive region with different settings of parameters $\gamma$ and $\lambda$. The comparison results of the accuracy are depicted in Fig.~\ref{fig:comparison_1} for methods of fixed region proposal, adaptive region proposal, WHT-only method and DCT method. In these comparisons, three cases of $\gamma$ values (case 1 of $\gamma = 8$, case 2 of $\gamma = 16$, and case 3 of $\gamma = 32$) are plotted with three iterations of 20, 35 and 50. It can be observed from results that, the adaptive region proposal and the fixed region proposal outperform the WHT-only method and the DCT method. It can be observed that, the adaptive region proposal has improved accuracy results compared with the fixed region proposal, the WHT-only method and the DCT method. The computation time comparisons are given in Table~\ref{comparison_computation_time}. 

\begin{table}[]
\small
\centering
\caption{Training computation time in seconds for the proposals and the methods in comparison under various parameters $\gamma$ and $\lambda$.}\label{comparison_computation_time}
\begin{tabular}{m{2.5cm}m{1.02cm}m{1.02cm}m{1.02cm}m{1.02cm}} 
\\ \hline
\textbf{Method}  & $\gamma$ = $8$, $\lambda$ = $100$ & $\gamma$ = $8$, $\lambda$ = $200$ & $\gamma$ = $16$, $\lambda$ = $500$ & $\gamma$ = $32$, $\lambda$ = $500$
\\ \hline \hline
Our fixed region  &  273  &  832   &   3,149  &   2,385
\\ \hline
Our adaptive region &  258	  &  813	   &  2,832   &    2,268
\\ \hline
WHT-only method   &  981  &  2,271  &  6,342   &  5,418
\\ \hline
DCT method &  1,462   &  2,359    &   6,518   &  4,517
\\ \hline
\end{tabular}
\vspace{-0.1in}
\end{table}

Note that, the sizes of image matrices after the region adaptation are zero-padded to be the same size so that the CNN structure can process these images. The threshold value of $\eta$ is adjusted to $0.997$ to obtain the results of the adaptive region proposal. The adaptive region proposal has achieved fastest computation speed in comparison with other methods. Therefore, the adaptive region proposal is the preferred method for this underwater object classification task. The benefits of reducing the training time and improving the accuracy performance are due to the noises in the underwater object images and the regional max-pooling operation in the proposal. The scattering effects of water and particles in the natural water cause the images to be blurry, lack of contrast and has color distortion. The two-stage WHT operation extracts major information of the images into the inner region, and removes the noisy information in the outer region. The max-pooling operation further filters the noises in the outer region and reduces the data size. Therefore, the proposal can achieve the training time reduction while maintaining the performance compared with the competing schemes. Furthermore, the adaptive region proposal has improved the performance in terms of accuracy compared with the fixed region proposal. The training time reduction is of more significant practical value compared with the inference time reduction. The underwater robots need to collect the images data to do the training of the CNN models then deploy the trained CNN model in the inference of the underwater objects. Therefore, it is expected that the training time is relatively short so that the underwater robots can proceed to underwater navigation shortly after the underwater image collection. Our proposal has reduced the training time while maintaining the image classification inference time, so the proposal is of practical value in underwater robotics.

\balance

\section{Conclusions}\label{conclusions}
We proposed new methods for source-feature compression based on the partitioning of the tensor after two-stage Walsh-Hadamard Transform for the application of underwater object classification. We introduced two novel proposals to partition the regions, i.e., a fixed-region and an adaptive-region proposal. Afterwards, the compressed data matrices were sent to the Convolutional Neural Network for classification. The proposals were evaluated with the experimentally collected underwater object datasets and verified to effectively reduce the computation time while improving the classification accuracy compared with the competing methods. The adaptive region proposal improves the performance compared with the fixed region proposal. Besides the object classification application, the proposed approaches have applications in computer vision tasks in learning-based underwater robotics including the underwater object detection and underwater image enhancement. The future applications of the proposal can also be expanded to the general image classification fields to classify public image datasets and the video sequence datasets.


\bibliographystyle{IEEEtran}
\bibliography{references_object_recognition.bib}

\begin{thebibliography}{10}
\providecommand{\url}[1]{#1}
\csname url@samestyle\endcsname
\providecommand{\newblock}{\relax}
\providecommand{\bibinfo}[2]{#2}
\providecommand{\BIBentrySTDinterwordspacing}{\spaceskip=0pt\relax}
\providecommand{\BIBentryALTinterwordstretchfactor}{4}
\providecommand{\BIBentryALTinterwordspacing}{\spaceskip=\fontdimen2\font plus
\BIBentryALTinterwordstretchfactor\fontdimen3\font minus
  \fontdimen4\font\relax}
\providecommand{\BIBforeignlanguage}[2]{{%
\expandafter\ifx\csname l@#1\endcsname\relax
\typeout{** WARNING: IEEEtran.bst: No hyphenation pattern has been}%
\typeout{** loaded for the language `#1'. Using the pattern for}%
\typeout{** the default language instead.}%
\else
\language=\csname l@#1\endcsname
\fi
#2}}
\providecommand{\BIBdecl}{\relax}
\BIBdecl

\bibitem{chen2019real}
W.~Chen, M.~Rahmati, V.~Sadhu, and D.~Pompili, ``Real-time image enhancement
  for vision-based autonomous underwater vehicle navigation in murky waters,''
  in \emph{Proceedings of the ACM International Conference on Underwater
  Networks and Systems (WUWNet)}, 2019, pp. 1--8.

\bibitem{Deveci18}
T.~C. Deveci, S.~{\c{C}}akir, and A.~E. {\c{C}}etin, ``Energy efficient
  {Hadamard} neural networks,'' \emph{CoRR}, vol. abs/1805.05421, 2018.

\bibitem{Verma18}
``{DCT}-domain deep convolutional neural networks for multiple {JPEG}
  compression classification,'' \emph{Signal Processing: Image Communication},
  vol.~67, pp. 22 -- 33, 2018.

\bibitem{Rahman18}
S.~Rahman, A.~Q. Li, and I.~M. Rekleitis, ``{SVIn2}: Sonar visual-inertial
  {SLAM} with loop closure for underwater navigation,'' \emph{CoRR}, vol.
  abs/1810.03200, 2018.

\bibitem{rahmatislam2018}
M.~Rahmati, S.~Karten, and D.~Pompili, ``{SLAM}-based underwater adaptive
  sampling using autonomous vehicles,'' in \emph{Proceedings of MTS/IEEE OCEANS
  Conference}, 2018, pp. 1--7.

\bibitem{lu2015contrast}
H.~Lu, Y.~Li, L.~Zhang, and S.~Serikawa, ``Contrast enhancement for images in
  turbid water,'' \emph{Journal of the Optical Society of America A (JOSA A)},
  vol.~32, no.~5, pp. 886--893, 2015.

\bibitem{Alom18}
M.~Z. Alom, T.~M. Taha, C.~Yakopcic, S.~Westberg, M.~Hasan, B.~C.~V. Esesn,
  A.~A.~S. Awwal, and V.~K. Asari, ``The history began from {AlexNet}: {A}
  comprehensive survey on deep learning approaches,'' \emph{CoRR}, vol.
  abs/1803.01164, 2018.

\bibitem{Alex12}
K.~Alex, I.~Sutskever, and G.~E. Hinton, ``{ImageNet} classification with deep
  convolutional neural networks,'' in \emph{Advances in Neural Information
  Processing Systems (NIPS)}, 2012, pp. 1097--1105.

\bibitem{Karen14}
K.~Simonyan and A.~Zisserman, ``Very deep convolutional networks for
  large-scale image recognition,'' \emph{CoRR}, vol. abs/1409.1556, 2014.

\bibitem{He16}
K.~{He}, X.~{Zhang}, S.~{Ren}, and J.~{Sun}, ``Deep residual learning for image
  recognition,'' in \emph{IEEE Conference on Computer Vision and Pattern
  Recognition (CVPR)}, 2016, pp. 770--778.

\bibitem{Szegedy15}
C.~{Szegedy}, {Wei Liu}, {Yangqing Jia}, P.~{Sermanet}, S.~{Reed},
  D.~{Anguelov}, D.~{Erhan}, V.~{Vanhoucke}, and A.~{Rabinovich}, ``Going
  deeper with convolutions,'' in \emph{IEEE Conference on Computer Vision and
  Pattern Recognition (CVPR)}, 2015, pp. 1--9.

\bibitem{Chollet16}
F.~Chollet, ``Xception: Deep learning with depthwise separable convolutions,''
  \emph{CoRR}, vol. abs/1610.02357, 2016.

\bibitem{Howard17}
A.~G. Howard, M.~Zhu, B.~Chen, D.~Kalenichenko, W.~Wang, T.~Weyand,
  M.~Andreetto, and H.~Adam, ``{MobileNets}: Efficient convolutional neural
  networks for mobile vision applications,'' \emph{CoRR}, vol. abs/1704.04861,
  2017.

\bibitem{Sandler18}
M.~Sandler, A.~G. Howard, M.~Zhu, A.~Zhmoginov, and L.~Chen, ``Inverted
  residuals and linear bottlenecks: Mobile networks for classification,
  detection and segmentation,'' \emph{CoRR}, vol. abs/1801.04381, 2018.

\bibitem{Aiyer05}
A.~Aiyer, K.~P. Pyun, Y.~zong Huang, D.~B. O’Brien, and R.~M. Gray, ``{Lloyd}
  clustering of {Gauss} mixture models for image compression and
  classification,'' \emph{Signal Processing: Image Communication}, vol.~20,
  no.~5, pp. 459 -- 485, 2005.

\bibitem{Davis19}
E.~K. Davis, U.~Majumder, C.~Capraro, C.~Cicotta, J.~Siddall, and D.~Brown,
  ``{An approaches for noise induced object classifications accuracy
  improvement},'' in \emph{Cyber Sensing}, I.~V. Ternovskiy and P.~Chin, Eds.,
  vol. 11011, 2019, pp. 41 -- 50.

\bibitem{Sanchez11}
J.~{Sanchez} and F.~{Perronnin}, ``High-dimensional signature compression for
  large-scale image classification,'' in \emph{IEEE Conference on Computer
  Vision and Pattern Recognition (CVPR)}, 2011, pp. 1665--1672.

\end{thebibliography}

\end{document}